%% file: main.tex
\input{_constants}
\arxiv

\documentclass[10pt,twocolumn,letterpaper]{article}
\input{cvpr_header}
\begin{document}
\maketitle
\begin{abstract}
Generated images of score-based models can suffer from errors in their spatial means, an effect, referred to as a color shift, which grows for larger images. This paper investigates a previously-introduced approach to mitigate color shifts in score-based diffusion models. We quantify the performance of a nonlinear bypass connection in the score network, designed to process the spatial mean of the input and to predict the mean of the score function. We show that this network architecture substantially improves the resulting quality of the generated images, and that this improvement is approximately independent of the size of the generated images. As a result, this modified architecture offers a simple solution for the color shift problem across image sizes. We additionally discuss the origin of color shifts in an idealized setting in order to motivate the approach.
\end{abstract}

\section{Introduction}
Score-based diffusion models approximate the score of a data distribution in order to generate synthetic data via a diffusion process. The data is first transformed into a latent space via a simple diffusion (or noising) process; the probability distribution of the state over the latent space is known. Sampling a generated image reverses the diffusion process, transforming noise (a sample from the latent space) into realistic images. In training, a loss function is optimized in order to learn an approximation of the score function \citep{Vincent}, and in the sampling, a differential equation is solved numerically which depends on the score function \citep{YangSongSDE}. Models of this type have been used successfully to generate realistic images, video, and audio (e.g., \citet{kong2020diffwave, HoVideoDiffusion, ho2022imagen}), and they have outperformed GANs on image synthesis tasks (e.g., \citet{DhariwalNichol2021, Palette}).

However, without specific adjustments to the network architecture or training implementation, these models can produce errors in the generated images referred to as ``color shifts" \citep{Song2020_Improved}.   In this work, we demonstrate that the color shift is primarily an error in the spatial mean of the generated images. We investigate a simple and effective method \citep{bischoff2023unpaired} for improving these color shifts in generated images. We demonstrate how this method works using the FashionMNIST dataset \citep{xiao2017fashionmnist} and using snapshots from a high-resolution dynamical simulation of two-dimensional forced turbulent fluid flow \citep{bischoff2023unpaired}. We contrast this method to other solutions in terms of ease of implementation and performance as a function of image size and dataset size. 

\subsection{Related Work}\label{sec:related_work}
In the original work identifying the color shift, it was observed that generated color images created with a ``vanilla" score-based diffusion model exhibit coherent shifts in color (e.g., entire images shifted to redder colors). The color shift was worse for larger images. When using the same training iteration model checkpoint, the generated samples had similar color shifts, but the color shift changed as training continued \citep{Song2020_Improved}. The authors proposed storing a version of the model parameters which update via an exponential-moving average (EMA) and using these parameters during sampling; this worked well to alleviate the color shift and improved the overall image quality for images ranging in size from 32x32 to 256x256. These observations are consistent with a model with high variance, at least with respect to the task of predicting the general color of an image.

Other authors have proposed modifying the loss function, network architecture, or sampling process in order to alleviate the color shift. Sampling an image using a diffusion model involves solving a differential equation from an initial condition (e.g., at $t=1$) to the final state (at $t=0$), such that the solution at $t=0$ is the generated image. Optimizing the loss function corresponds to learning the tendency function (the derivative of the image with respect to time) \textit{at all times} from $t=0$ to $t=1$. Large-scale spatial features are generated earlier (nearer to $t=1$) in the differential equation integration (assuming that the large scale features have a larger amplitude than small scale features), and \citet{Choi2022} suggested that weighting those times more in the loss function could lead to improved image quality/reduced color shifts. Modifying the sampling process to include a projection onto the training data manifold also improved the color shift and image quality \citep{mei2022bi}. However, both of these works used EMA-smoothed models, and so it is not clear how much the weighting or sampling modifications additionally contributed to reducing the color shift.  \citet{ProgressiveDistillation} note that the objective of the score network can be framed in different ways: predicting the noise added to an image at time $t$, predicting the noised image itself, and predicting a linear combination of the the two; \citet{ho2022imagen} found that using the latter objective prevented color shifts in larger images.   Finally, attention layers are used in many score network architectures \citep{Ho2020, YangSongSDE}, and possibly also alleviate the color shift since they allow for nonlocal learning in CNNs \citep{nonlocalnn}.

Separately, there has been work connecting color shifts directly to errors in the spatial mean of the generated images. \citet{wang2023exploiting} found that re-normalizing the spatial means of the samples with the mean and variance of the spatial means in the true data alleviated the color shift for their use case. \citet{bischoff2023unpaired} employed an alternative network architecture, consisting of a bypass layer for predicting the spatial means of images, and a U-net \citep{unet} for predicting spatial variations about the mean. They noted that this network architecture alleviated color shifts in generative modeling of two-dimensional turbulent fluid flow. No additional investigations were performed to demonstrate the performance of the mean-bypass layer as a function of image size, to compare it to other color-shift correction methods, or to apply it to other datasets.

\subsection{Our contribution}
We identify the color shift primarily as an error in the spatial means of images. Coherent shifts in color which preserve a realistic spatial structure, as observed in the original work observing the color shift \citep{Song2020_Improved}, are consistent with shifts in the spatial mean in one or more of the channels (this implies that ``color shift" errors can also occur in grayscale images). We show how errors in the spatial means of the generated images must result from errors in the predicted score's spatial mean. This suggests that CNNs trained with a score-based diffusion loss function struggle to accurately model spatial means in the score, even though this is a vastly simpler task compared with predicting the spatial variations about the mean. 

To that end, we employ the same neural architecture as \citep{bischoff2023unpaired}, consisting of a 2-layer feed-forward network which predicts the spatial mean of the score (the ``mean-bypass layer"), and a standard U-net \citep{unet} which handles the component of the score involving variations about the spatial mean.  The loss function is approximated such that it has two independent terms, one for the mean score, and one for the score which predicts spatial variations.  This turns the initial problem of using a single network with a single loss to predict the entire score function into one which uses two sub-networks with non-overlapping parameter sets, that are trained simultaneously with two independent loss functions, and that are used simultaneously in sampling. We do not use attention layers or weighting schemes. 

Our primary contributions are to motivate the mean-bypass layer of \citep{bischoff2023unpaired} mathematically, and to compare, using multiple datasets, the performance of models which employ this architecture against a baseline which employs EMA to address the color shift. 

\section{Mathematical motivation}\label{sec:theory}
Consider the forward noising process in the variance exploding case of score-based generative models (e.g., \citet{Ho2020, YangSongSDE}),
\begin{equation}\label{eq:forward_sde}
    \dd \vec{x} = g(t) \dd \vec{V}_t,
\end{equation}
where $\vec{V}_t$ is a Wiener process and the standard deviation of the noising process is determined by the prescribed function $g(t)$, resulting in a Brownian motion without drift. The associated reverse diffusion process is given by
\begin{equation}\label{eq:rev_sde}
    \dd{\vec{x}} = -g(t)^2 \vec{s}(\vec{x},t) \dd{t} + g(t) \dd{\vec{W}_t},
\end{equation}
where $\vec{W}_t$ is again a Wiener process and $\vec{s}(\vec{x},t)$ is the gradient of the log of the probability distribution $p(x,t)$ that satisfies the Fokker-Planck equation corresponding to \Cref{eq:forward_sde} (\citet{ANDERSON1982}, Equations 5.5-5.7).

We can carry out a Reynolds decomposition of $\vec{x}$, writing $\vec{x}(t) = \vec{x}'(t) + \bar{\vec{x}}(t)$, where $\bar{\vec{x}}$ is the spatial mean of $\vec{x}$ and $\vec{x}'$ denotes a zero mean field of spatial variation about the mean. Then without loss of generality, we can separate the reverse diffusion process given by \Cref{eq:rev_sde} into two coupled reverse processes
\begin{align}
    \dd{\vec{\bar{x}}} &= -g(t)^2 \vec{\bar{s}}(\vec{x}, t) \dd{t} + g(t) \dd{\vec{\overline{W}}_t} \label{eq:reynolds1_mean}\\
    \dd{\vec{x'}} &= -g(t)^2 \vec{s}'(\vec{x}, t) \dd{t} + g(t) \dd{\vec{W}'_t}, \label{eq:reynolds1_spatial}
\end{align}
one for the spatial mean, and one for the spatially varying part of $\vec{x}$. At this stage, it is already clear that errors in the spatial mean of the score control errors in the spatial mean of the generated image (the solution at $t=0$ to \Cref{eq:reynolds1_mean}). Changes in the mean shift all pixel values of the full data sample $\vec{x}$, leading to errors in the colors of color images and darkening or whitening in grayscale images. This error may also affect the spatially part of $\vec{x}$ via the term $\vec{s}'(\vec{x}, t)$.

To further motivate the network design of \citep{bischoff2023unpaired}, we make a simplifying assumption for the sake of illustration: that the spatial mean and the spatially varying component of $\vec{x}$ are independent random variables
\begin{equation}\label{eq:p_spatial_mean}
    p(\vec{x}, t) = p(\bar{\vec{x}}, \vec{x}', t) = p(\bar{\vec{x}},t) p(\vec{x}',t).
\end{equation}
If this were true, we can think of sampling from $p(\bar{\vec{x}},t)$ and $p(\vec{x}',t)$ separately, and further assume that
\begin{align}\label{eq:split_score}
    \vec{\bar{s}}(\vec{x}, t) &\approx \vec{\bar{s}}(\bar{\vec{x}}, t)\\
    \vec{s}'(\vec{x}, t) &\approx \vec{s}'(\vec{x}', t).
\end{align}
This simplifies Equations~\eqref{eq:reynolds1_mean} and \eqref{eq:reynolds1_spatial} to yield two approximately independent reverse diffusion processes
\begin{align}\label{eq:reynolds2}
    \dd{\vec{\bar{x}}} &= -g(t)^2 \vec{\bar{s}}(\bar{\vec{x}}, t) \dd{t} + g(t) \dd{\vec{\overline{W}}_t}\\
    \dd{\vec{x'}} &= -g(t)^2 \vec{s}'(\vec{x}', t) \dd{t} + g(t) \dd{\vec{W}'_t}. 
\end{align}
This shows how, when we solve the reverse SDE to sample images, we can think of approximately solving a system of uncoupled SDEs, i.e., sampling an image mean and the spatial variations about the mean, and then adding them together to obtain $\vec{x}$. 

While it may be possible to argue that these assumptions might approximately hold for very large images (approximate de-correlation of means and spatial structure), we do \emph{not} assume that it holds in general. We only use this motivation to design our network architecture to predict spatial means and spatial variation about the mean independently. In practice, this means only that correlations between the mean and the spatial structure will not be used in training.

\section{Score network architecture design}

When employing the variance exploding form of the forward noising process, the noised image at time $t$, 
$\vec{x}(t)$, is drawn from a normal distribution $\mathcal{N}(\vec{x}(0), \sigma^2(t)\mathcal{I})$ with a known $\sigma(t)$ (see e.g., \citet{Ho2020, YangSongSDE} for details).  The score function is modeled using a neural network $\vec{f}_\theta$ as
\begin{equation}\label{eq:score_total}
    \vec{s}_\theta(\vec{x},t) = \frac{\vec{f}_\theta(\vec{x},t)}{\sigma(t)},
\end{equation}
which allows $\vec{f}_\theta$ to target a value of order unity \citep{Song2020_Improved}. We can rewrite this as
\begin{equation}\label{eq:score_split}
    \vec{s}_\theta(\vec{x},t) = \bar{\vec{s}}_\theta(\vec{x},t) + \vec{s}'_\theta(\vec{x},t)  = \frac{\bar{\vec{f}}_\theta(\vec{x},t)}{\sigma(t)} + \frac{\vec{f}'_\theta(\vec{x},t)}{\sigma(t)},
\end{equation}
where again the overbar indicates a spatial mean, and the prime (e.g., $\vec{f}' = \vec{f}-\bar{\vec{f}}$) indicates spatial variations about the mean. Given this representation of the score function, the denoising score-matching loss function introduced in \citep{Ho2020, YangSongSDE} can be written as \citep{bischoff2023unpaired}
\begin{align}\label{eq:loss_general}
\mathcal{L}(\theta) \approx \mathcal{L}'(\theta) + \bar{\mathcal{L}}(\theta) &= \mathbb{E}_{t, \vec{x}(0), \vec{x}(t)} \bigg[ (\vec{f}'_\vec{\theta}(\vec{x},t) + \vec{\epsilon}')^2 \bigg] \nonumber \\
&+  \mathbb{E}_{t, \vec{x}(0), \vec{x}(t)} \bigg[ (\bar{\vec{f}}_\vec{\theta}(\vec{x},t) + \bar{\vec{\epsilon}})^2   \bigg],
\end{align}
where
\begin{equation}
\mathbb{E}_{t, \vec{x}(0), \vec{x}(t)} =  \mathbb{E}_{t\sim U(0,1], \vec{x}(0) \sim p(\vec{x}(0)), \vec{x}(t) \sim p(\vec{x}(t) | \vec{x}(0)) },
\end{equation}
and $\vec{\epsilon} \sim \mathcal{N}(\vec{0},\mathcal{I})$ as a Gaussian random variable. In deriving this, the expectation of the quantity $\vec{f}' \bar{\vec{f}}$ is assumed to be zero, which is reasonable given that they are approximating uncorrelated quantities (i.e., because the expectation of $\vec{\epsilon}' \bar{\vec{\epsilon}}$ is zero). 

Note that $\bar{\vec{\epsilon}}$ must be $\mathcal{O}(1/N)$, where $N^2$ is the total number of pixels in an image. We can therefore factor a term proportional to $1/N^2$ from the second loss term $\bar{\mathcal{L}}$. 
This suggests a possible remedy for color shifts: introduce a weighting factor in front of $\overline{\mathcal{L}}$ to emphasize the loss for the spatial mean. However, we carried out experiments with different weightings of this term, and found that this does not fix the color shift, even in the extreme case of \textit{only} including $\overline{\mathcal{L}}$ in the loss function. A reason for this is suggested if we instead rewrite this loss term in terms of a sum over independent Fourier modes instead of over spatial and mean components. In that case, we would see that each mode has a factor of $1/N^2$, and there is nothing special about the magnitude of the ``mean loss". This suggests that there is an inductive bias in the baseline score network architecture against learning the spatial mean of the score. 

\subsection{Baseline score network}\label{sec:baseline}
The baseline network we compare against is a U-net \citep{unet}, which maps $\vec{x}$ and $t$ to the predicted score via $\vec{f}_\theta$ as in \Cref{eq:score_total}. We use three downsampling layers, followed by 8 residual blocks which preserve the size and number of channels, followed by three upsampling layers. Initial lifting and final projection layers are also employed. All convolutional kernels are of size 3x3. The time variable is embedded using a Gaussian Fourier embedding \citep{tancik2020fourier}. The parameters of the U-net are updated via gradient descent. An EMA-smoothed model is also computed and used to generate the images in some of our results when noted.

\subsection{Modified score network}\label{sec:arch}
Using a single network to learn both $\bar{\vec{s}}$ and $\vec{s}'$ via optimizing the loss of \Cref{eq:loss_general} can yield errors in the generated image means; we have shown that this can only arise from errors in $\bar{\vec{s}}$. An alternative approach is to instead turn the problem into two independent tasks with independent models and loss functions, but to train them simultaneously.  

Following \citep{bischoff2023unpaired}, the first action of $\vec{f}_\theta$ is to split $\vec{x}$ into a spatial mean $\bar{\vec{x}}$ and variations about the mean $\vec{x}'$. The part of the score network which predicts the spatial variations about the mean score, $\vec{s}'$, is the same U-net architecture described in \Cref{sec:baseline}, which here maps two inputs ($\vec{x}'$ and $t$), to a single output. The mean is removed from this output to produce the prediction. If we denote the parameters of the U-net by $\phi$, and the U-net function by $\vec{u}$, we have
\begin{equation}\label{eq:spatial_f_net}
\vec{f}'_\phi(\vec{x}',t) = \vec{u}_\phi(\vec{x}',t) - \overline{\vec{u}'_\phi(\vec{x}',t)}.
\end{equation}
The part of the score network which predicts the spatial mean in the score is a dense feed-forward network, referred to as the mean-bypass layer. If we denote the parameters of this mean-bypass layer by $\Phi$, and the network function itself by $\bar{\vec{n}}$, we have
\begin{equation}\label{eq:mean_f_net}
\bar{\vec{f}}_\Phi(\bar{\vec{x}},t) = \frac{\bar{\vec{n}}_\Phi(\bar{\vec{x}},t)}{N},
\end{equation}
where we have multiplied the output of the mean bypass layer by a factor of $1/N$. This allows the function $\bar{\vec{f}}_\Phi$ to produce a quantity of order unity - in the loss function, \Cref{eq:loss_general}, it is seeking to match a quantity of order $\mathcal{O}(1/N)$. We found that not including this factor led to poorer performance.

The score function is obtained by plugging in \Cref{eq:spatial_f_net} and \Cref{eq:mean_f_net} into \Cref{eq:score_split}, where $\theta$ has been split into two non-overlapping sets $[\phi, \Phi]$. Plugging Equations~\eqref{eq:spatial_f_net} and \eqref{eq:mean_f_net}) into \Cref{eq:loss_general}, we obtain the following expression for the components of the score matching loss
\begin{align}\label{eq:loss_split}
\mathcal{L}'(\phi) &= \mathbb{E}_{t, \vec{x}(0), \vec{x}(t)} \bigg[ (\vec{f}'_\vec{\phi}(\vec{x}',t) + \vec{\epsilon}')^2 \bigg]\\
\bar{\mathcal{L}}(\Phi) &= N^{-2}\mathbb{E}_{t, \vec{x}(0), \vec{x}(t)} \bigg[ (\bar{\vec{n}}_\vec{\Phi}(\bar{\vec{x}},t) + \bar{\vec{\epsilon}}^*)^2   \bigg], 
\end{align}
where $\bar{\vec{\epsilon}}^*$ is a standard normal random variable. Training to optimize this loss, using the architecture described in \Cref{sec:arch}, yields two independent optimization problems, one for the spatially-varying part and one for the spatial mean part of $\vec{f}$. As with the baseline model, an exponentially-averaged version of the modified model is used to generate the images in some of our results as noted. 

\section{Experiments}\label{sec:experiments}
In order to investigate the performance of the modified network, we conduct a series of experiments with two datasets. For both, we train and compare the performance of the baseline model described in \Cref{sec:baseline} and modified model with the mean-bypass layer described in \Cref{sec:arch}, both with and without EMA. 

Our first dataset is FashionMNIST \citep{xiao2017fashionmnist}, which has around 60,000 data points with a resolution of 28x28 and is available with an MIT license. We create interpolated versions at resolutions of 32x32, 64x64, 128x128, 256x256, and 512x512 pixels. This dataset is used to explore the color shift magnitude as a function of image size. The second dataset consists of snapshots from a two-dimensional fluid flow simulation. The raw training data consists of 800 samples of 512x512 images with multiple channels corresponding to different dynamical state variables. We focus here on the moisture supersaturation field (used to compute precipitation); results for the vorticity were similar. Further details regarding the dataset are given in \citet{bischoff2023unpaired}. This dataset is used to demonstrate the performance of the modified network compared to the baseline network on natively high-resolution images and to demonstrate that the color shift is primarily an error in image means.

We used an Adam optimizer \citep{kingma2014adam} with a learning rate of 0.001. Other optimizer hyperparameter values, as well as the EMA rate of 0.999, were taken from \citep{YangSongSDE}. The batch size was dictated by memory limits, and the number of training epochs was determined using loss curves and model performance.  All our samples are generated with an Euler-Maruyama timestepper. The commands to reproduce the training and sampling runs which generate the FashionMNIST results are provided in the README of the code repository \hyperlink{https://anonymous.4open.science/r/Easing-Color-Shifts-5473}{https://anonymous.4open.science/r/Easing-Color-Shifts-5473}.

\begin{figure*}
  \centering
  \includegraphics[width=\textwidth]{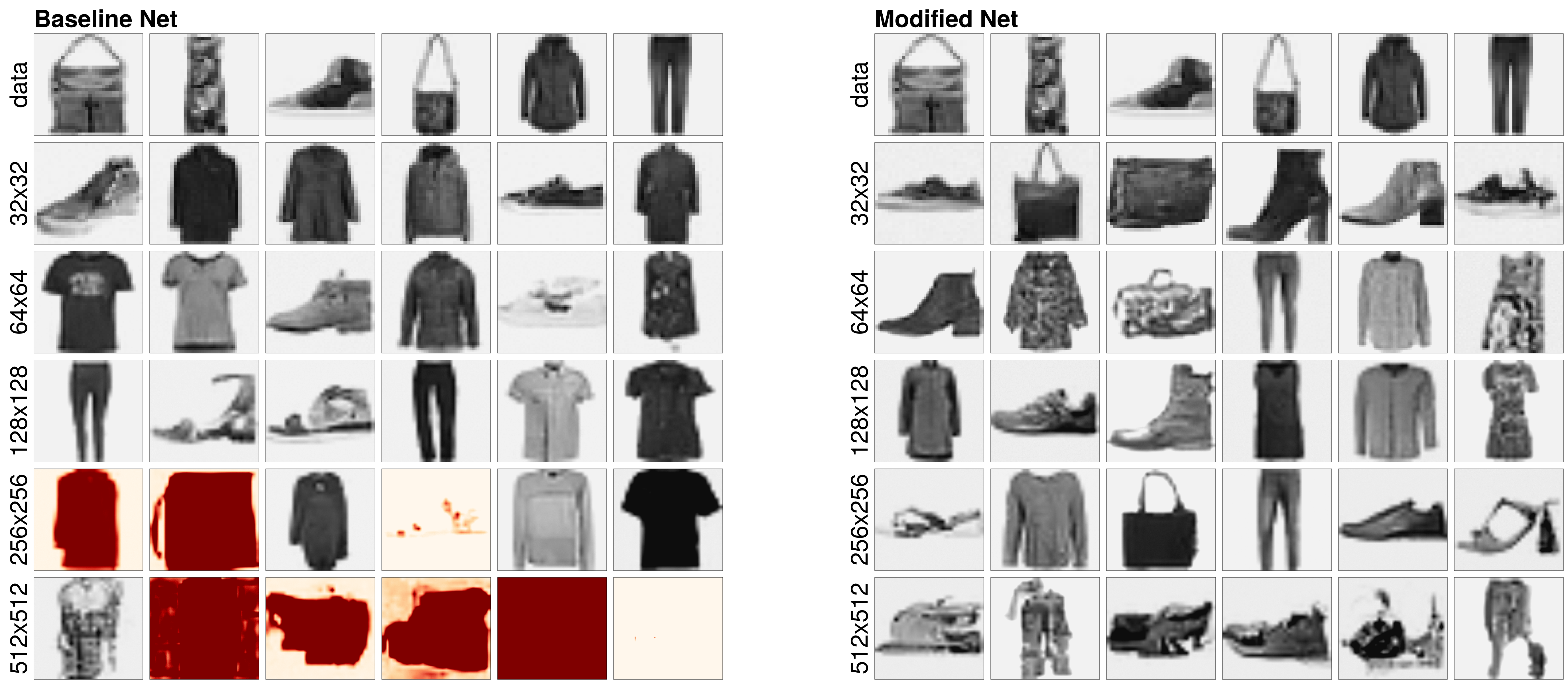}
  \caption{\textbf{Data samples} generated using a baseline U-net approach (left columns) and with our modifications (right columns). The top rows shows random data samples from FashionMNIST \citep{xiao2017fashionmnist}. Each subsequent row shows random generated samples for the two model at different FashionMNIST resolutions. The samples colored in \textit{red} are those with strong color shifts. Both models used EMA to smooth parameters; comparisons between the two sets of results are therefore reflective of what our modifications contribute in terms of performance.}
  \label{fig_examples}
\end{figure*}
\subsection{FashionMNIST}
All FashionMNIST experiments use the same training hyperparameters, including a batch size of $16$ and $100$ training epochs (or $\sim1$e5 training iterations, which is comparable to the number used in \citep{Song2020_Improved}). We trained the models and sampled images on a single NVIDIA V100 GPU; it took roughly 48 hours to train the model with 256x256 images for 100 epochs. For both the baseline and the modified experiments, we used EMA-parameters to generate samples; comparisons between the two sets of results are therefore reflective of what the mean-bypass layer contributes in terms of performance.

We first demonstrate the effect of training with the baseline model. Samples drawn with the baseline U-net are shown in \Cref{fig_examples} (left columns). While the baseline U-net, which employs only EMA as a color shift correction, performs well with the smaller resolution datasets, the color shift becomes worse as the image size is increased. On the other hand, when the mean-bypass layer is active and the spatial mean is processed separately, the color shift is negligible for all resolutions (right columns).

\begin{figure*}
  \centering
  \includegraphics[width=\textwidth]{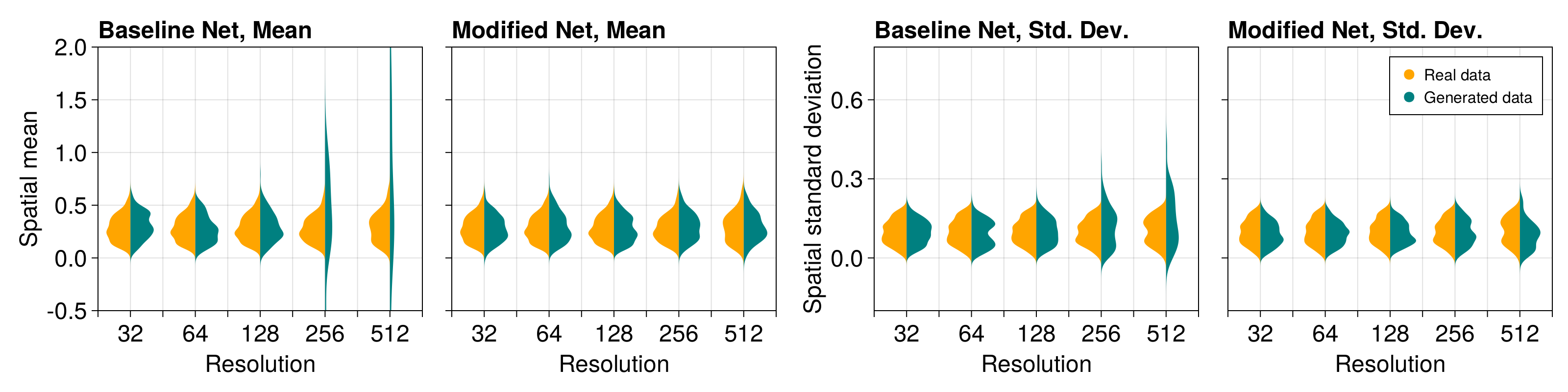}
  \caption{Violin plots of \textbf{spatial means} and \textbf{spatial standard deviations}. The left and third-to-left plots show results for the baseline U-net at different resolutions of FashionMNIST \citep{xiao2017fashionmnist}. The second-to-left and rightmost plots show results for the U-net with mean-bypass layer. The \textit{yellow} distributions show the training data, while the \textit{green} distributions show the corresponding quantity generated by the reverse diffusion processes. All distributions are based on kernel density estimates of 200 data samples, except for the 256x256 and 512x512 cases, where we use 100 and 33 data samples, respectively. Both models used EMA to smooth parameters; comparisons between the two sets of results are therefore reflective of what our modifications contribute in terms of performance.}  
  \label{fig_means_and_stds}
\end{figure*}

\Cref{fig_means_and_stds} shows the distributions of spatial means and spatial standard deviations for different resolutions of the FashionMNIST dataset. The left and third-to-left plots show the results for the baseline U-net, demonstrating that the distributions of generated spatial mean and standard deviation differ increasingly from the true underlying data as the resolution of the image is increased, but that the spatial mean has a much larger error. This implies that using an exponential moving average alone is not enough to alleviate the color shift in large images. This is also consistent with our overall thesis that images exhibiting a color shift can still retain approximately correct spatial structure, i.e. that color shifts are predominantly errors in spatial means. However, the error in the spatial standard deviation still increases with image size. This may be related to \Cref{eq:reynolds1_spatial}: for the baseline model, where a single function approximates $\vec{s}(\vec{x}) = \vec{s}'(\vec{x}) + \bar{\vec{s}}(\vec{x})$, large enough errors in the spatial mean eventually must propagate into the score $\vec{s}'$ via $\vec{x}$.   The second-to-left and rightmost plots in \Cref{fig_means_and_stds} show the results for the U-net with mean-bypass layer, showing improved results for the spatial mean (and standard deviation) distributions when compared with the true data, regardless of image size.

\begin{figure*}
  \centering
  \includegraphics[width=\textwidth]{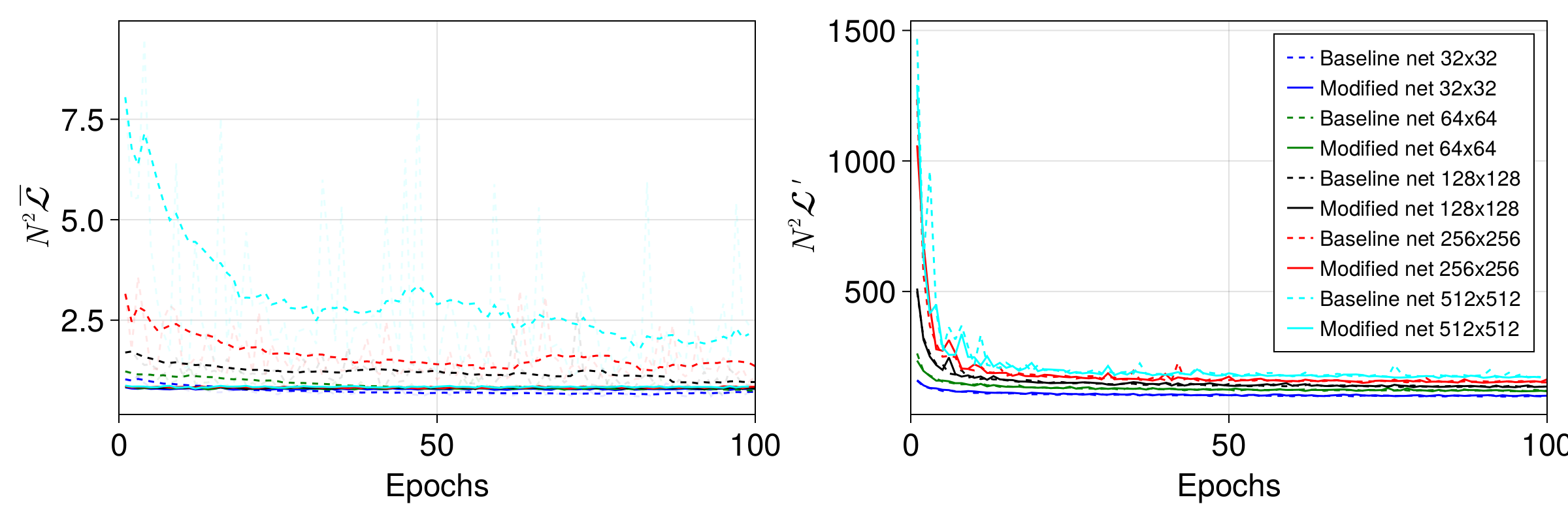}
  \caption{\textbf{Loss curves} for train and test errors (no EMA applied). The left panel shows the loss curves for the spatial mean loss $\overline{\mathcal{L}}$, while the right panel shows the loss curves for the spatially varying loss $\mathcal{L}'$. $N$ is the horizontal number of pixels. In the left panel, we smoothed the loss curves for the baseline net to make them easier to see. The original data is still shown but with increased transparency. It can be seen that while the spatially varying loss reaches low values quickly, the spatial mean loss takes time to reach low values for the baseline model. This problem is absent with the modified network.}
    \label{fig_loss_curves}
\end{figure*}

\begin{figure*}
  \centering
  \includegraphics[width=\textwidth]{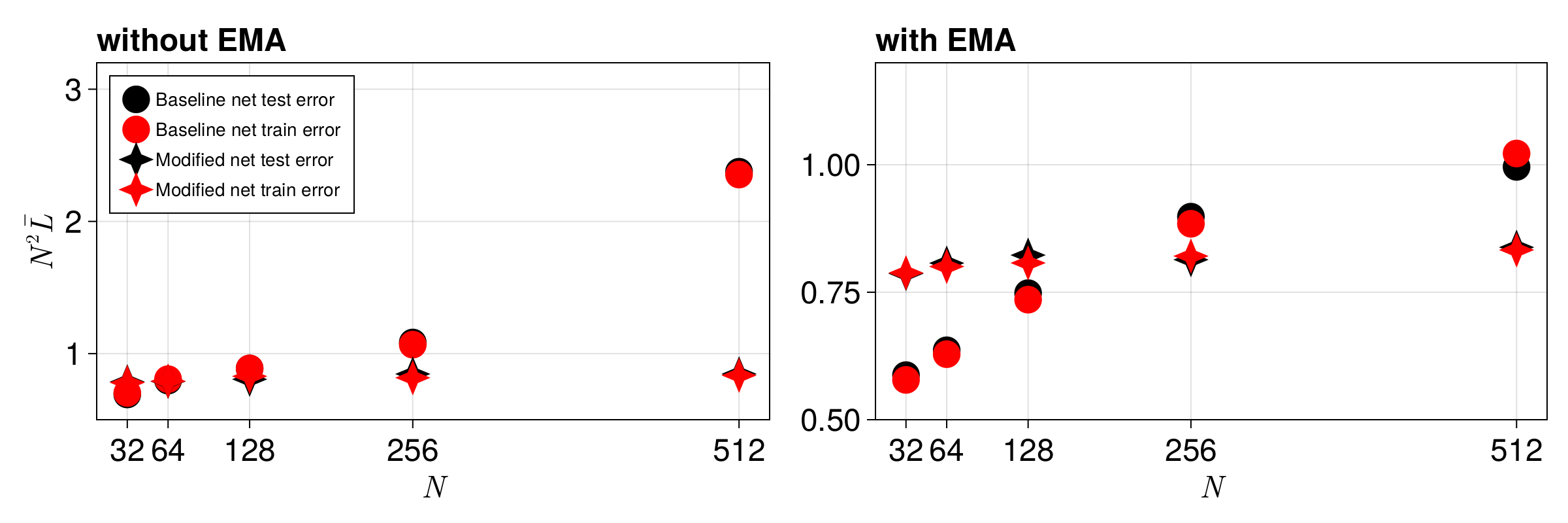}
  \caption{\textbf{Spatial mean loss} after 100 epochs of training for different resolutions of FashionMNIST \citep{xiao2017fashionmnist}, with (right) and without (left) EMA applied. $N$ is the horizontal number of pixels. The spatial mean error for the baseline network increases with resolution, while the spatial mean error for the modified network stays nearly flat with increasing resolution.}
    \label{fig_losses_weighted}
\end{figure*}
The training curves for the per-pixel losses $N^2\bar{\mathcal{L}}$ and $N^2\mathcal{L}'$ for all image sizes and for both models (\Cref{eq:loss_split}) are shown in \Cref{fig_loss_curves}. A comparison of the right panel, for the spatial loss $N^2\mathcal{L}'$, and the left panel, for the mean loss $N^2\bar{\mathcal{L}}$, show that, when using the baseline model, the spatial loss is optimized comparatively quickly.  A comparison of the dotted and solid curves in the left panel show that, when using the modified model with the mean-bypass layer, the spatial mean is learned more rapidly, regardless of image size, and without high variance, compared to the baseline model. Note that in these figures, we did not use the smoothed model.

We can make the observation of how well the term $N^2\bar{\mathcal{L}}$ is optimized as a function of image size more concrete. In \Cref{fig_losses_weighted}, we plot the value of this loss at the end of training using the unsmoothed models (on the left) and the smoothed models (on the right). When using the mean-bypass layer modification, the final loss grows weakly with image size. This is true regardless of if EMA is used (right) or not (left), i.e. using EMA does not materially improve the performance of the modified network in terms of optimizing the spatial loss term.  On the other hand, the baseline model, without the mean-bypass layer, has a loss value which grows more rapidly as a function of image size. This growth in error with image size occurs regardless of whether EMA is used to smooth the parameters (right), or not (left), though using EMA reduces the overall value of the loss achieved by the baseline model significantly.

Poorer optimization of the spatial loss manifests as an error in the mean score, which in turn manifests as an error in the mean of the generated images. For large enough images, these results suggest that smoothing of parameters will not be enough to remove the color shift.
\subsection{Two-dimensional turbulence}
\begin{figure*}
  \centering
  \includegraphics[width=\textwidth]{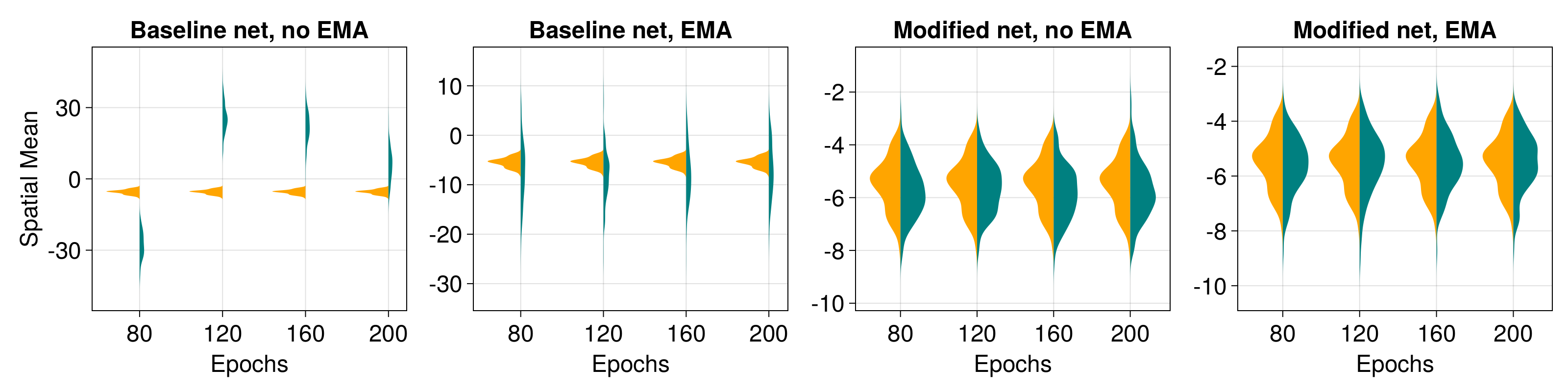}
  \caption{\textbf{Spatial mean distributions} resulting from different model configurations as noted. The \textit{yellow} distributions show the spatial means of the training data, while the \textit{green} distributions show the corresponding spatial means generated by the reverse diffusion process using a model from different training epochs. All distributions are based on kernel density estimates of 100 data samples. The modified network shows increased performance over the baseline network for higher resolutions, regardless of whether EMA is used.}
  \label{fig_2dturb_means}
\end{figure*}
Both 2D turbulence experiments use the same training hyperparameters, including a batch size of $4$, $200$ training epochs, and a dropout probability of 0.5. The number of training epochs implies $\sim4$e4 training iterations, which is 5-10x smaller than the number used in \citep{Song2020_Improved}. All figures in this section show the results for the moisture supersaturation variable.

In \Cref{fig_2dturb_means}, we show the distribution of spatial means from generated images and from the data in four cases: with the baseline model, with the smoothed baseline model, with the modified model, and with the smoothed modified model. We find that the color-shift is present when using the baseline net without EMA, and that the mean of the distribution is corrected when using EMA with the baseline net, but not the variance. Since EMA parameters retain some memory of the past training steps, we expected to see the smoothed baseline model converging over training epoch. That EMA does not alleviate the color shift in this example may be because we did not train for enough iterations; regardless, the modified network is able to predict spatial means correctly with this amount of training. We also see that the modified network improves the color shift regardless of if EMA is used or not. 

\begin{figure*}
  \centering
  \includegraphics[width=\textwidth]{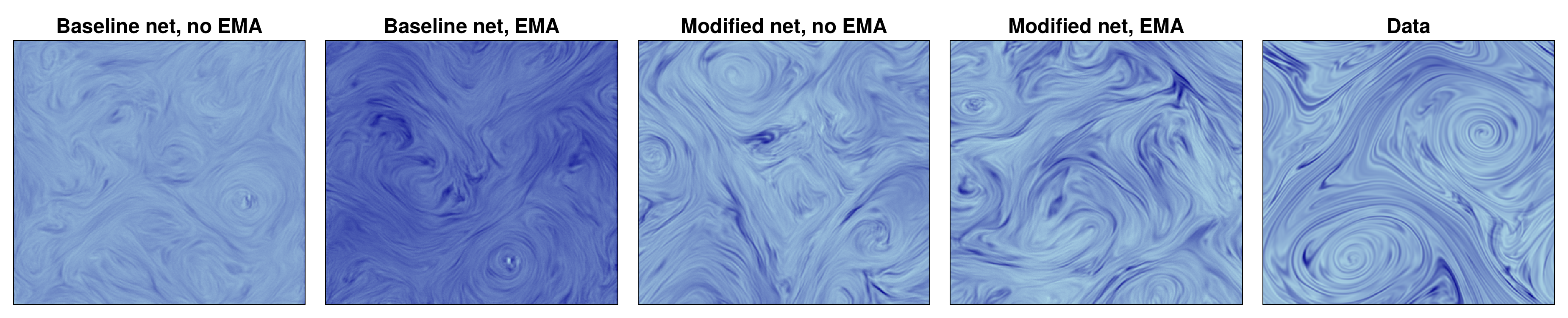}
  \caption{\textbf{De-meaned data samples} generated using different model configurations as noted (first four images, left to right) and taken from the data (right). All images use the same colorbar limits.}
  \label{fig_2dturbexamples}
\end{figure*}

\begin{figure*}
  \centering
  \includegraphics[width=\textwidth]{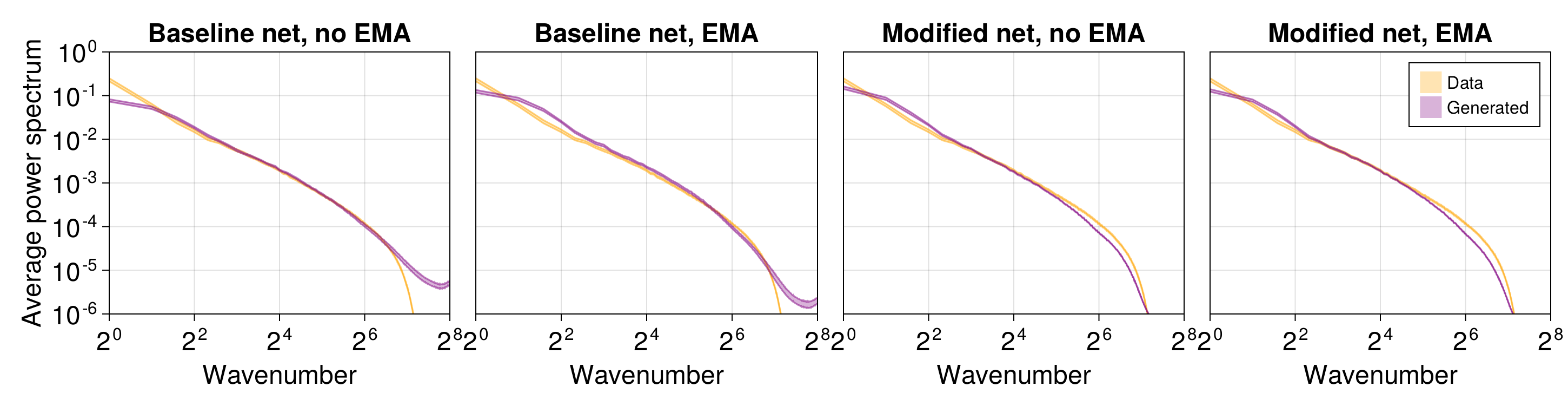}
  \caption{\textbf{Azimuthally averaged power spectra} computed from images using different model configurations as noted. Yellow spectra show spectra of real samples and purple spectra show spectra of generated samples. Shaded areas are computed from 100 bootstrap samples at the 90\% confidence interval.}
  \label{fig_spectra_ch1}
\end{figure*}

In \Cref{fig_2dturbexamples}, we show de-meaned generated images, in addition to a de-meaned data sample. These were generated using the models from epoch 200. Qualitatively, the spatial structure is being learned by all four networks. To further investigate how well the spatial structure is modeled, we computed the azimuthally averaged power spectrum of 100 samples generated by each of the four models and taken from the data. These are shown in \Cref{fig_spectra_ch1}. Even the baseline network without EMA generates images that have a power spectrum that differs from that of the data by at most a factor of a few. This suggests that errors in spatial means are somewhat decoupled from errors in spatial structure, which supports the choice of network architecture and the identification of a color shift with errors in spatial means.

\section{Discussion}\label{sec:conclusions}
In this article, we have investigated color shifts in images generated by score-based diffusion models. We began by presenting a mathematical argument connecting errors in the spatial means of generated images, which result in color shifts in color images, to errors in the spatial mean of the score function. By examining the value of different components of the loss function, we demonstrated that a baseline network, with no corrections for the color shift, learns the spatial mean of the score more slowly than it learns the significantly more complex objective of learning spatial variations about the mean score. Similarly, by examining the power spectra of de-meaned images, we showed that the baseline network does learn spatial structure even when the spatial means are poorly predicted. 

Our mathematical argument and these empirical results naturally motivate the use of the spatial mean bypass architecture \citep{bischoff2023unpaired} as a solution to the color shift, because it allows for generating means with a separate model. Using different datasets, we demonstrated that the same-sized mean-bypass layer removes the color shift almost independently of the size of the original images, but that the performance of the baseline network with EMA degrades more rapidly with image size. For the largest images studied (512x512), we found that EMA alone is not enough to alleviate the color shift over the training period. The bypass layer does not materially change the complexity of the standard U-net network or the number of free parameters, and it does not require tuning any weighting schedule within the loss or additional hyperparameters. It provides an appealing alternative solution to the color shift, especially for large images.

\subsection{Limitations}
In motivating the mean-bypass layer architecture, we used a Reynolds decomposition of the reverse SDE, which by itself is general (Equations \eqref{eq:reynolds1_mean} and \eqref{eq:reynolds1_spatial}). However, in our actual implementation, we only pass the image mean $\bar{\vec{x}}$ to the approximator for $\bar{\vec{s}}$, and we only pass the variations about the mean $\vec{x}'$ into the approximator for $\vec{s}'$. If correlations are present in the data between image means and the spatial variations, our architecture does not make use of them.  While it would be trivial to pass the entire image to the U-net, to partially remove this limitation, it would require more design to provide information about $\vec{x}'$ to the mean-bypass layer.
\section*{Acknowledgments}
The research presented in this work has been supported by Eric and Wendy Schmidt (by recommendations of the Schmidt Futures) and by the Cisco Foundation. We thank Tapio Schneider for insightful discussions on this work. All calculations were performed with the help of Caltech's Resnick High Performance Computing Center.

{
    \small
    \bibliographystyle{ieeenat_fullname}
    \bibliography{main}
}
\end{document}

%% file: _constants.tex
\def\paperID{10227} 
\def\confName{CVPR}
\def\confYear{2024}

\title{Easing Color Shifts in Score-Based Diffusion Models}

\author{Katherine Deck\\
Climate Modeling Alliance\\
California Institute of Technology\\
{\tt\small kdeck@caltech.edu}
\and
Tobias Bischoff\thanks{Authors are listed in reverse alphabetical order. All authors contributed equally to this work.}\\
Ford Model e \\
{\tt\small tobias.bischoff@pm.me}
}

\newif\ifreview 
\newif\ifarxiv \newcommand{\arxiv}{\arxivtrue}
\newif\ifcamera 
\newif\ifrebuttal 

%% file: cvpr_header.tex
\ifreview \usepackage[review]{cvpr} \fi
\ifarxiv \usepackage[pagenumbers]{cvpr} \fi
\ifrebuttal \usepackage[rebuttal]{cvpr} \fi
\ifcamera \usepackage{cvpr} \fi

\input{_macros}  

\usepackage{xr-hyper}

\makeatletter
\newcommand*{\addFileDependency}[1]{
  \typeout{(#1)}
  \@addtofilelist{#1}
  \IfFileExists{#1}{}{\typeout{No file #1.}}
}

\makeatother

\definecolor{cvprblue}{rgb}{0.21,0.49,0.74}
\usepackage[pagebackref,breaklinks,colorlinks,citecolor=cvprblue]{hyperref}
\usepackage[capitalize]{cleveref}
\crefname{section}{Sec.}{Secs.}
\crefname{table}{Table}{Tables}
\crefname{figure}{Fig.}{Figs.}

%% file: _macros.tex
\usepackage{times}
\usepackage{graphicx}	
\usepackage{amsmath}	
\usepackage{amssymb}	
\usepackage{booktabs}
\usepackage{lmodern}
\usepackage{microtype}
\usepackage{epsfig}
\usepackage[table,xcdraw,dvipsnames]{xcolor}
\usepackage{caption}
\usepackage{float}
\usepackage{placeins}
\usepackage{color, colortbl}
\usepackage{stfloats}
\usepackage{enumitem}
\usepackage{tabularx}
\usepackage{xstring}
\usepackage{multirow}
\usepackage{xspace}
\usepackage{url}
\usepackage{subcaption}
\usepackage{xcolor}
\usepackage[hang,flushmargin]{footmisc}

\ifcamera \usepackage[accsupp]{axessibility} \fi

% ...

\ifarxiv  \fi

\newcommand{\R}[1]{{%
    \textbf{%
        \ifstrequal{#1}{1}{\textcolor{red}{R#1}}{%
        \ifstrequal{#1}{2}{\textcolor{blue}{R#1}}{%
        \ifstrequal{#1}{3}{\textcolor{magenta}{R#1}}{%
        \ifstrequal{#1}{4}{\textcolor{teal}{R#1}}{%
                           \textcolor{cyan}{R#1}%
        }}}}%
    }%
}}

\newcommand{\dd}[1]{\mathrm{d}#1}

\renewcommand{\vec}[1]{\mathbf{\boldsymbol{#1}}}